\documentclass[a4paper]{article}

\usepackage{INTERSPEECH2019}
\usepackage{enumitem}
\usepackage{amsmath}
\usepackage{subcaption}
\usepackage{ulem}
\usepackage{siunitx}
\usepackage{lipsum}
\usepackage{graphicx}

\usepackage[dvipsnames]{xcolor}

\usepackage{cite}
\usepackage{microtype}

\usepackage{booktabs}
\usepackage{tabularx}
\newcolumntype{C}{>{\centering\arraybackslash}X}
\newcolumntype{L}{>{\raggedright\arraybackslash}X}
\newcolumntype{R}{>{\raggedleft\arraybackslash}X}
\newcolumntype{P}[1]{>{\raggedright\arraybackslash}p{#1}}

\newcommand{\func}[2]{#1_\text{#2}}

\title{On the Contributions of Visual and Textual Supervision \\ in Low-Resource Semantic Speech Retrieval}
\name{Ankita Pasad$^1$, Bowen Shi$^1$, Herman Kamper$^2$, Karen Livescu$^1$}
\address{
  $^1$Toyota Technological Institute at Chicago, USA\\
  $^2$Dept.\ E\&E Engineering, Stellenbosch University, South Africa}
\email{\{ankitap,bshi,klivescu\}@ttic.edu, kamperh@sun.ac.za}

\begin{document}

\maketitle
\begin{abstract}
Recent work has shown that speech paired with images can be used to learn semantically meaningful speech representations even without any textual supervision. In real-world low-resource settings, however, we often have access to some transcribed speech. We study whether and how visual grounding is useful in the presence of varying amounts of textual supervision. In particular, we consider the task of semantic speech retrieval in a low-resource setting. We use a previously studied data set and task, where models are trained on images with spoken captions and evaluated on human judgments of semantic relevance. We propose a multitask learning approach to leverage both visual and textual modalities, with visual supervision in the form of keyword probabilities from an external tagger. We find that visual grounding is helpful even in the presence of textual supervision, and we analyze this effect over a range of sizes of transcribed data sets. With $\sim$5 hours of transcribed speech, we obtain 23\% higher average precision when also using visual supervision.
\end{abstract}

\section{Introduction}\label{sec:intro}
In many languages and domains, there can be insufficient data to train modern large-scale models for common speech processing tasks. Recent work has begun exploring alternative sources of weak supervision, such as visual grounding from images~\cite{synnaeve+etal_nipsworkshop14,harwath+etal_nips16,chrupala+etal_acl17,chrupala2018symbolic,scharenborg+etal_icassp18,kamper+roth_sltu18,kamper+etal_taslp19}. Such work has developed approaches for training models for tasks such as cross-modal retrieval~\cite{harwath+etal_nips16,chrupala+etal_acl17,chrupala2018symbolic}; unsupervised learning of word-like and phone-like  units~\cite{gelderloos+chrupala_coling16,harwath+glass_acl17,harwath2018jointly,harwath2019towards}; and retrieval tasks like keyword search, query-by-example search, and semantic search~\cite{kamper+etal_interspeech17,kamper+etal_taslp19,kamper+etal_icassp19}. Much of this recent work is in the context of {\it zero} textual supervision, that is, using no transcribed speech, and the results show that visual grounding alone provides a strong supervisory signal.  However, in many low-resource settings we also have access to a small amount of textual supervision. This is the setting that we address here. In particular, we consider the task of semantic speech retrieval using models trained on a data set of images paired with spoken and written captions.

In semantic speech retrieval, the input is a textual query{, a written word in our setting,} and the task is to find utterances in a speech corpus that are semantically relevant to the query~\cite{garofolo+etal_cbmia00,lee+etal_taslp15}. The query word need not exactly occur in the retrieved utterances; for example, the query {\it beach} should retrieve the exactly matched utterance ``a dog retrieves a branch from a {\it beach}" {as well as} the semantically matched utterance ``people at an oceanside resort." One common approach is to cascade an automatic speech recognition (ASR) system with a text-based information retrieval method~\cite{chelba+etal_ieee08}. High-quality ASR, however, requires significant amounts of transcribed speech audio and text for language modelling. Kamper {\it et al.}~\cite{kamper+etal_interspeech17,kamper+etal_taslp19} proposed learning a semantic speech retrieval model solely from images of natural scenes paired with unlabelled spoken captions. This approach uses an external visual tagger to automatically obtain text labels from the training images, and these are used as targets for training a speech-to-keyword network. No textual supervision is thus used. For analysis, the approach was compared to a supervised model trained only with text supervision. The visually supervised model was shown to have an edge over the textually trained model in retrieving non-exact keyword matches. In retrieving the exact matches, the text-based model proved superior. 

Based on these observations, we hypothesize that visual- and text-based supervision can be complementary for semantic speech retrieval in low-resource settings where both are available. Visual supervision could, e.g., provide a signal to distinguish acoustically similar but semantically unrelated words. Here we consider a regime where the amount of transcribed speech audio is not enough to train a full ASR system, but is enough to provide a useful additional supervision signal. Using a corpus of images with spoken captions, we propose a multitask learning (MTL) approach where visual supervision is obtained from an external image tagger and text supervision is obtained from  (a limited amount of) transcribed spoken captions. Each type of target is given as a bag of words. Additionally, we  include a representation learning loss to encourage similarity between the visual and spoken representations.

We consider different amounts of transcribed speech audio to investigate the effect of MTL in different low-resource settings and to understand the trade-off between visual and textual supervision. Experiments using human semantic relevance judgments show that our new MTL approach incorporating both modalities performs better than using either one in isolation.We achieve consistent improvements over the entire range of transcribed data set sizes considered. This demonstrates that the benefit of visual grounding remains even when textual supervision is available.
  
\begin{figure*}
        \centering
        \includegraphics[width=0.8\textwidth,height=5.6cm]{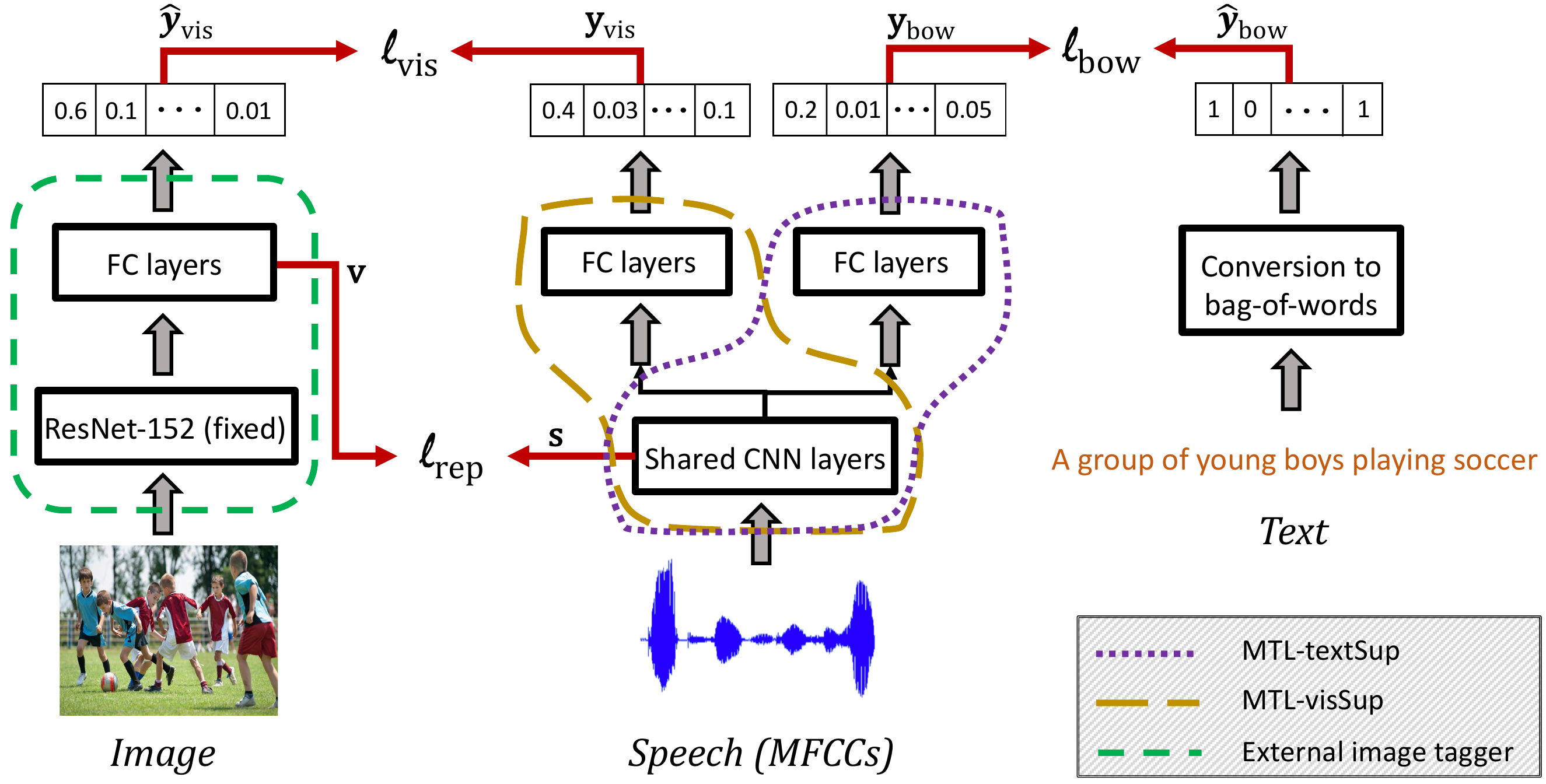}
        \caption{The multitask speech model with visual supervision (left) and textual supervision (right). Either of the textually supervised and visually supervised branches (or both, ensembled) can be used during inference.}
        \label{fig:model}
        \vspace*{-10pt}
\end{figure*}

\section{Approach}\label{sec:framework}
    
As a starting point, we use the model of~\cite{kamper+etal_taslp19}, which addresses the semantic retrieval task using a training set of images and their spoken captions with no textual supervision at all. In this approach, a {\it vision model}---an image tagger trained on an external data set of tagged images---is used to produce a bag of semantic keywords for each image, along with their posterior probabilities. These posteriors serve as supervision for a {\it speech model} that takes as input a spoken utterance and outputs a bag of keywords describing the utterance. At test time, these predicted keywords are used for semantic speech retrieval, by retrieving those utterances for which the model predicts a probability higher than a given {\it detection threshold} for an input query word.
    
We make two key changes to the model of~\cite{kamper+etal_taslp19}, as illustrated in Figure \ref{fig:model}. First, we consider the case where we have both the visual supervision (the vector of keyword probabilities) and textual supervision, obtained from transcriptions of the training utterances. Specifically, for the textual supervision, we convert each transcription into a bag of ground-truth content words occurring in the utterance. The textual and visual supervision are used together in a multitask learning (MTL) approach, where our model consists of two branches, one that produces the ``visual keywords'' and one that produces ``textual keywords''.
    
The second change we make is to add a {\it representation loss} term. The purpose of this is to encourage intermediate representations learned by the speech model to match those in the external visual tagger. This loss is similar to ones used in recent work on unsupervised joint learning of visual and speech representations, e.g.\ for cross-modal retrieval~\cite{harwath+etal_nips16}.  In our case the representation loss can be viewed as a regularizer or as a third task in the MTL framework. This loss helps us take advantage of the fact that the visual tagger is trained on much more data than the speech model, so we expect its internal representations themselves to be useful during training of the speech network.
    
As a final, minor change, we use a stronger pre-trained visual network (ResNet-152) in the visual tagger relative to~\cite{kamper+etal_taslp19} (more details about the vision model are given in Section~\ref{exp}).
    
We note that other concurrent work also investigates MTL for speech with visual and textual supervision (applied to different tasks) \cite{chrupala2018symbolic}. A key distinction is that we focus on the specific question of how useful visual grounding is in the presence of varying amounts of textual supervision.

\subsection{Model Details} \label{sec:opti}
Each training utterance $U={\bm u}_1, {\bm u}_2,...,{\bm u}_T$ is paired with an image $I$. Each frame ${\bm u}_t$ is an acoustic feature vector, MFCCs in our case.\footnote{Earlier work on this data set compared MFCCs to filterbank features, and found that MFCCs worked similarly or better~\cite{kamper+etal_interspeech17}.} The vision model provides weak labels, $\func{\bm y}{vis}\in [0,1]^{\func{N}{vis}}$, where $\func{N}{vis}$ is the number of keywords in the visual tag set. This serves as the ground truth for the visually supervised branch output, $\func{f}{vis}(U) = \func{\hat{\bm y}}{vis}$, of the speech model.

Each training utterance $U$ is also optionally paired with a multi-hot bag-of-words vector  $\func{\bm y}{bow}\in\{0,1\}^{\func{N}{bow}}$, obtained from the transcriptions of the spoken captions, where $\func{N}{bow}$ is the number of unique keywords in the text labels and each dimension indicates the presence of absence of a particular word. This vector serves as a ground truth for the output of the textually supervised branch, $\func{f}{bow}(U) = \func{\hat{\bm y}}{bow}$, of the speech model.

These task-specific supervised losses are optimized jointly with an unsupervised multi-view representation loss. In particular, we use a margin-based contrastive loss~\cite{hermann+blunsom_iclr14} between the speech representation $\bm s$ and visual representation $\bm v$ at an intermediate layer within each model. 

The total loss is a weighted sum of these three losses: $ \ell = \func{\alpha}{vis}\func{\ell}{vis} + \func{\alpha}{bow}\func{\ell}{bow} + (1-\func{\alpha}{bow}-\func{\alpha}{vis})\func{\ell}{rep}$, where each of the loss terms is a function of a training utterance $U$ and either (1) a corresponding image representation ${\bm v}$, (2) a visual target ${\bm y}_\textrm{vis}$, or (3) a textual target ${\bm y}_\textrm{bow}$ (if available). Both of the supervised task losses, $\func{\ell}{vis}$ and $\func{\ell}{bow}$, are summed cross entropy losses between the predicted and ground-truth vectors, as follows (where sup$\in$\{vis,bow\}):
\begin{equation*}
\vspace*{-0.2cm}
    \func{\ell}{sup} 
    = -\sum_{w=1}^{|\func{N}{sup}|} \left\{\func{y}{sup,w} \log \func{\hat{y}}{sup,w} + (1-\func{y}{sup,w}) \log [1-\func{\hat{y}}{sup,w}] \right\} \label{eq:ce}
\vspace*{0.2cm}
\end{equation*}
where $\func{\hat{\bm y}}{sup}$ is a function of the speech input and $\func{\bm y}{sup}$ is function of the image/transcription. The representation loss is a contrastive loss, similar to ones used in prior work on multi-view representation learning~\cite{hermann+blunsom_iclr14,harwath+etal_nips16,he2017multi}, computed by sampling a fixed number of negative examples within a mini-batch (size $B$) corresponding to both images and utterances: 
\vspace{-0.18cm}
\begin{align*}
    \func{\ell}{rep} 
    &= \left\{ \frac{1}{|V|}\sum_{\bm v'\in V}\max[0,m+ \func{d}{cos}(\bm v,\bm s)-\func{d}{cos}(\bm v',\bm s)] \right. \nonumber \\[-4pt]
    &\ \ \left.+ \frac{1}{|S|} \sum_{\bm s' \in S}\max[0,m+\func{d}{cos}(\bm v,\bm s)-\func{d}{cos}(\bm v,\bm s')] \right\} \label{eq:contrastive}
\end{align*}
where $\{\bm v,\bm s\}$ are the representations of the correct vision-speech pair; 
$\{\bm v',\bm s\}$ and $\{\bm v,\bm s'\}$ are negative (non-matching) pairs;
$\func{d}{cos}(\bm v,\bm s)$ is the cosine distance between the representations;
$\func{n}{neg} = |V| = |S|$ is the number of negative pairs; and $m$ is a margin indicating the minimum desired difference between the  positive-pair distances and negative-pair distances.

\section{Experimental Setup}\label{exp}
\subsection{Data}\label{sec:data}
  We use {three training sets and one evaluation data set}, identical to those used in prior work~\cite{kamper+etal_taslp19}, allowing for direct comparison. 
  
  {\it (A) Image-text pairs}: The union of MSCOCO \cite{coco} and Flickr30k \cite{flickr30k}, with $\sim$149k images ($\sim$107k training, $\sim$42k dev) paired with 5 written captions each. This is an external data set used to train the image tagger before our MTL approach is applied.  We are taking advantage of the fact that, in contrast to speech, labelled resources for images are more plentiful (in fact, we are implicitly also using the even larger training set of the pre-trained ResNet).  The availability of such large visual data sets allows us to train a strong external image tagger. 
  
  {\it (B) Image-speech pairs}: {The Flickr8k Audio Captions Corpus \cite{harwath+glass_asru15}}, consisting of $\sim$8k images paired with 5 spoken captions each, amounting to a total of $\sim$46 hours of speech data ($\sim$34 hours training, $\sim$6 hours dev, and $\sim$6 hours test speech data).  The images in this set are disjoint from those in set A.
  
  {\it (C) Speech-text pairs}: The spoken captions in the Flickr8k Audio Captions Corpus have written transcripts as well. We use subsets of these transcripts with varying sizes: from just $\sim$21 minutes to the complete $\sim$34 hours of labelled speech. 
    
  {\it (D) Human semantic relevance judgments}: For semantic speech retrieval evaluation, we use the human relevance judgments from~\cite{kamper+etal_taslp19}. 1000 utterances from the Flickr8k Audio Captions Corpus were manually annotated via Amazon Mechanical Turk with their semantic relevance for each of 67 query words. Each (utterance, keyword) pair was labeled by 5 annotators.  We use both the majority vote of the annotators (as ``hard labels") and the actual number of votes (``soft labels") for evaluation. 
  
  \subsection{Implementation Details} \label{sec:model}
 The {\it vision model} (Figure \ref{fig:model}, left) is an ImageNet pre-trained  ResNet-152 \cite{resnet} (which is kept fixed), with a set of four 2048-unit fully connected layers added at the top, followed by a final softmax layer that produces posteriors for the $\func{N}{vis}$ tags. This image tagger is used to provide the visual supervision ${\bm y}_\textrm{vis}$.  The fully connected layers in the vision model are trained once on set A and kept fixed throughout the experiments. 

 The {\it speech model} (Figure \ref{fig:model}, middle) consists of two branches, one with textual supervision and the other with visual supervision as targets. Except for the output layer, the architecture for these two networks is identical and exactly the same as the model in \cite{kamper+etal_taslp19}. Each network consists of three convolutional layers, with the output max-pooled over time to get a 1024-dimensional embedding, followed by feedforward layers, with a final sigmoid layer producing either $\func{N}{vis}$ or $\func{N}{bow}$ scores in $[0,1]$. The parameters for convolutional layers in these two branches are shared while the upper layer parameters are task-specific. Hereafter, the visually supervised branch and the textually supervised branch of this model are referred to as MTL-visSup and MTL-textSup respectively.
 The input speech is represented as MFCCs, zero-padded (or truncated\footnote{99.5\% of the utterances
are 8 s or shorter.}) to 8 seconds (800 frames).

For {\it representation learning} we need the dimensions for the learned speech and vision feature vectors to match. A two-layer ReLU feedforward network is used to transform the intermediate 2048-dimensional vision feature vectors to 1024 dimensions. The output of the shared branch of the speech model is used as the speech representation. Parameters of the additional layers are learned jointly with the speech model.

For MTL, each mini-batch is sampled from either set B or set C with probability proportional to the number of data points in each set, as in \cite{mini-batch}. The size of set C varies, as described in~\ref{sec:data}. $\func{N}{vis}$ is kept fixed at 1000 and $\func{N}{bow}$ is one of 1k, 2k, 5k and 10k (each time keeping the most common content words in set A) and optimized along with other hyperparameters. Hyperparameters include $\func{\alpha}{vis}$, $\func{\alpha}{bow}$, $m$, $\func{n}{neg}$, $\func{N}{bow}$, batch size, and learning rate. Different settings are found to be optimal as the set C is varied. Adam optimization~\cite{adam} is used for both branches with early stopping. For early stopping, we use $F$-score on the Flickr8k Audio Captions validation set with a detection threshold of 0.3. Since no development set is available for semantic retrieval, the $F$-score is evaluated for the task of exact retrieval (i.e., keyword spotting) using the 67 keywords from set D. 

\subsection{Baselines} \label{sec:baseline}
We compare our proposed models to two baselines. The {\it visual baseline} has access to visual supervision alone. This baseline replicates the vision-speech model from \cite{kamper+etal_taslp19}, but is improved here by using a pre-trained ResNet-152 (rather than VGG-16) for fair comparison with our models. The visual baseline is equivalent to MTL-visSup, when no textual supervision is used. The {\it textual baseline} has access to just the textual supervision, and is equivalent to MTL-textSup when using no visual supervision. Different baseline scores are obtained as the set C changes.

\subsection{Evaluation Metrics}\label{sec:eval}
At test time we have access to just the spoken utterances. For evaluation we use the output probability vectors of both, MTL-visSup and MTL-textSup. We evaluate semantic retrieval performance on set D (67 query words, 1000 spoken utterance search set). We measure performance using the following evaluation metrics, commonly used for retrieval tasks \cite{hazen+etal_asru09,zhang+glass_asru09}. {\it Precision at 10 ($P@10$)} and {\it precision at $N$ ($P@N$)} measure the precision of the top 10 and top $N$ retrievals, respectively, where $N$ is the number of ground-truth matches. {\it Average precision} is the area under the precision-recall curve as the detection threshold is varied. {\it Spearman's rank correlation coefficient} (Spearman's $\rho$) measures the correlation between the utterance ranking induced by the predicted probability vectors and the ground-truth ``probability'' vectors. The latter is approximated using the number of votes each query word gets for a given utterance. All of these metrics except for Spearman's $\rho$ are used for the hyperparameter tuning on exact keyword retrieval performance. 
    
    \section{Results and Discussion}\label{results}
\begin{figure}[!t]
    \centering
    \begin{subfigure}{\columnwidth}
        \includegraphics[width=\linewidth,height=5.5cm]{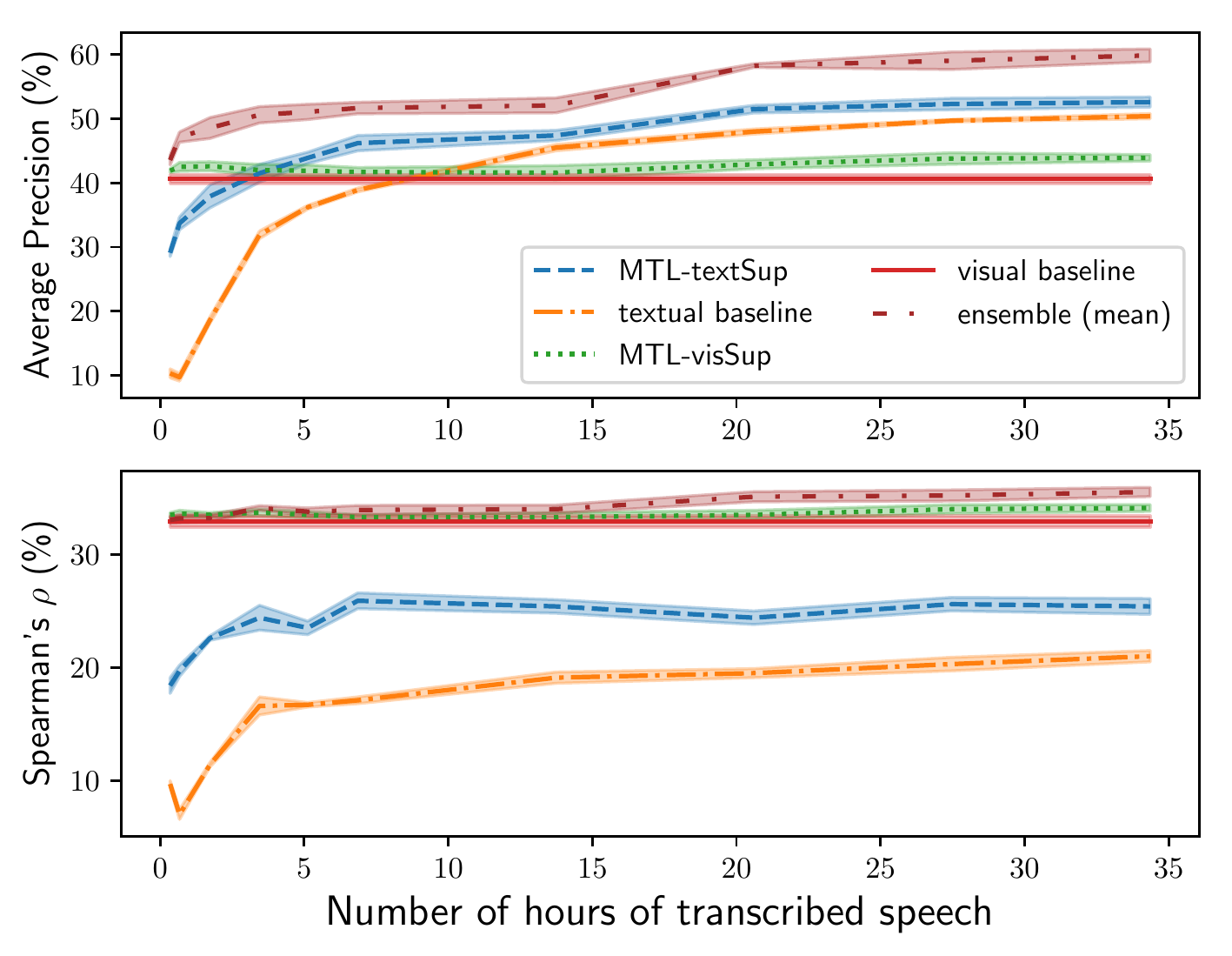}
    \end{subfigure}

    \vspace{15pt}
    
    \begin{subfigure}{\columnwidth}
        \centering
\vspace{-0.3cm}
    \centering
    \renewcommand{\arraystretch}{1.0}
    \scriptsize
   
  \begin{tabularx}{0.85\linewidth}{S[table-format=2.1] ccc}
        \toprule
        Hours & Textual baseline & Visual baseline & MTL ensemble \\
        \midrule
        1.7 & 18.6 & 40.6 & 48.6 \\
        34.4 & 50.4 & 40.6 & 59.9 \\
        \bottomrule
    \end{tabularx}

    \end{subfigure}

        \caption{Top: Semantic retrieval performance, in terms of average precision (AP) and Spearman's $\rho$.  Shading around the curves in the plots indicates standard deviation over multiple runs with different random initializations. Bottom: Average precision (\%) for a low-resource and higher-resource setting. }
        \label{fig:results-parallel}
\vspace{-0.35cm}
\end{figure}

Figure \ref{fig:results-parallel} presents the semantic retrieval performance of the MTL model as the amount of text supervision is varied.  For our MTL models, we can use either output (MTL-visSup or MTL-textSup) or the average of the two (ensemble). Note that the visual baseline result
is a horizontal line in each plot, as this baseline does not use any textual supervision. We present the results for Spearman's $\rho$, which evaluates with respect to the ``soft" human labels, and average precision, which uses the hard  majority decisions. The trends for other hard label-based metrics ($P$@10, $P$@$N$) follow the same trend as average precision. 

The first clear conclusion from these results is that visual grounding is still helpful even when some speech transcripts are available. This is, to our knowledge, the first time that this has been demonstrated for visually grounded models of speech semantics. In terms of average precision, the multitask model with both visual and textual supervision is better than the baseline text-supervised model even when using all of the transcribed speech. Ensembling the outputs gives a large performance boost over using only MTL-textSup: $\sim$28\% when using 1.7 hours (5\%) of transcribed speech and $\sim$14\% when using 34.4 hours (100\%). Even individually, the best performance at any level of supervision is obtained with one of the MTL models (MTL-textSup or MTL-visSup). 

In terms of Spearman's $\rho$, the trends are somewhat different from average precision.  Here, the MTL-visSup outperforms MTL-textSup at all levels of supervision, and the visual baseline alone is almost as good as the MTL-visSup.  This finding is in line with earlier work~\cite{kamper+etal_taslp19}, which found that the visual baseline does particularly well in terms of Spearman's $\rho$, which is more permissive of non-exact semantic matches.  In some sense, Spearman's $\rho$ uses a more ``complete" measure of the ground truth, since it considers the full range of human judgments rather than just the majority opinion. 

\vspace{0.06cm}
\noindent{\bf Effect of representation loss.} The results in Figure~\ref{fig:results-parallel} use the supervised losses as well as the representation loss.  Removing the latter reduces average precision by roughly 2-4\% for most points on the curve, while for Spearman's $\rho$ it makes little difference.  However, when tuning for exact retrieval on the development set, we find much larger gains from the representation loss, roughly 7-15\% in average precision.  This difference could be an artifact of tuning and testing on different tasks.

\vspace{0.06cm}
\noindent{\bf Effect of varying $\func{N}{bow}$.} We observe that a higher $\func{N}{bow}$ of 10k words is preferred at lower supervision, while at higher supervision having just 1k words in the output is best.  Our interpretation is that having more words in the output than we will be evaluating {on is equivalent to training the model on additional tasks which we do not care about at test time. This additional MTL-like setting has} a regularization effect which is helpful in lower-resource settings, but as we have access to more and more labelled data, we no longer benefit from regularization.

\vspace{0.06cm}
\noindent{\bf Other modeling alternatives.}  
We also considered other ways to combine the multiple sources of supervision: (1) by splitting the convolutional layers, (2) by pre-training on the visual task and fine-tuning on the textual task, or (3) by using a hierarchical multitask approach---inspired by prior related work~\cite{sogaard2016deep,toshniwal2017mtl,sanabria2018hierarchical} where the (more semantic) visual supervision is at a higher level and the (exact) textual supervision is at a lower level. Our MTL model (Figure~\ref{fig:model}) outperforms these other approaches.  

\begin{figure}[!t]
    \centering
    \includegraphics[width=\linewidth]{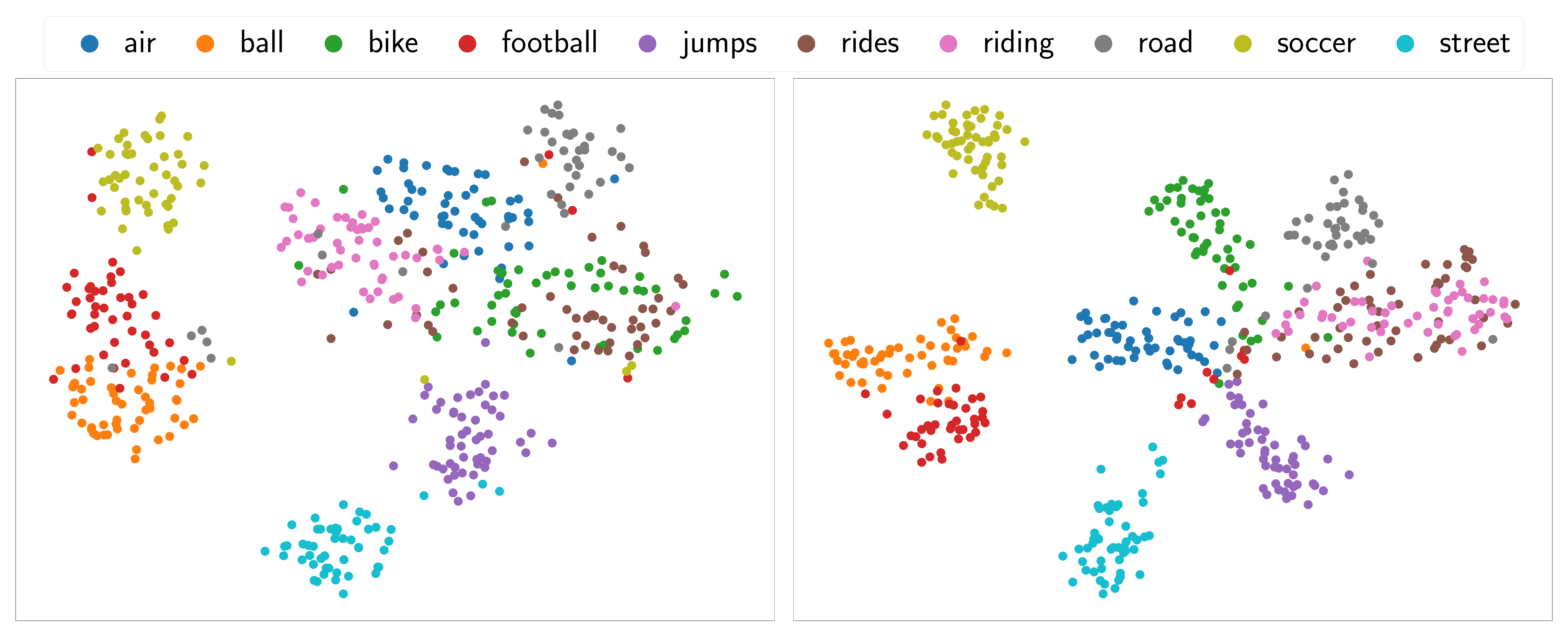}
    \caption{t-SNE visualization of the representations learned with $\sim$ 1.7 hours of transcribed speech, using the textual baseline (left) and MTL-textSup (right).}
    \label{fig:tsne}
    \vspace{-0.35cm}
\end{figure}

\vspace{-0.03in}
\noindent{\bf Qualitative analysis.} We visualize embeddings from the penultimate layer of the model by passing a set of isolated word segments through the model. Figure \ref{fig:tsne} shows 2D t-SNE~\cite{maaten2008visualizing} visualizations of these embeddings from both the textual baseline model and the MTL-textSup at low supervision.Qualitatively, the representation learned by the MTL model results in more distinct clusters than the baseline.We have also examined the retrieved utterances for these models and found that, at lower supervision, MTL-textSup outputs higher false positives than MTL-visSup due to acoustically similar words. For instance, for the query ``tree", MTL-textSup retrieves utterances containing the acoustically similar word ``street". This effect reduces as we train MTL-textSup on higher amount of text data. 

\vspace{-0.015in}
    \section{Conclusion}\label{sec:conclusion}
    
Visual grounding has become a commonly used source of weak supervision for speech in the absence of textual supervision. Our motivation here was to, first, examine the contribution of visual grounding when some transcribed speech is also available and, second, explore how best to combine both visual and textual supervision for a semantic speech retrieval task. We proposed a multitask speech-to-keyword model that has both visually supervised and textually supervised branches, as well as an additional explicit speech-vision representation loss as a regularizer.  We explored the performance of this model in a low-resource setting with various amounts of textual supervision, from none at all to 34 hours of transcribed speech. Our main finding is that visual grounding is indeed helpful even in the presence of textual supervision. Experiments over a range of levels of supervision show that joint training with both visual and textual supervision results in consistently improved retrieval.

A limitation of the current work is that the set of queries and human judgments is small.  A natural next step is to collect more human evaluation data and to consider a wider variety of queries, including multi-word queries.  Another next step is to widen the types of speech domains that we consider, for example to explore whether visually grounded training can learn retrieval models that are applicable also to speech that is not describing a visual scene. On a technical level, there is room for more exploration of different types of multi-view representation losses (such as ones based on canonical correlation analysis~\cite{ccae,gong2014multi}), as well as more structured speech models that can localize the relevant words/phrases for a given query. 
\vspace{-0.015in}
\section{Acknowledgements}
This material is based upon work supported by the Air Force Office of Scientific Research under award number FA9550-18-1-0166, by NSF award number 1816627, and by a Google Faculty Award to Herman Kamper.

    \newpage
    \bibliographystyle{IEEEtran}

    \bibliography{mybib}
    
\end{document}